\documentclass[letterpaper, 10 pt, conference]{ieeeconf}  %

\IEEEoverridecommandlockouts                              %

\usepackage[pdftex]{graphicx}
\usepackage{subcaption}
\usepackage{booktabs}

\usepackage{multirow}
\usepackage{amsmath}
\usepackage{amssymb}
\usepackage{amsfonts}
\usepackage{dsfont}
\usepackage[usenames,dvipsnames,svgnames,table,xcdraw]{xcolor}
\usepackage[colorlinks=true,pdfpagemode=UseNone,citecolor=black,linkcolor=black,urlcolor=BrickRed]{hyperref}
\usepackage{cleveref} %
\usepackage{algpseudocode}
\usepackage{algorithm}
\newcommand*\rot{\rotatebox{30}}

\usepackage{graphicx}
\usepackage{subcaption}
\usepackage{tikz}
\usetikzlibrary{matrix}
\usetikzlibrary{plotmarks, positioning}

\definecolor{myRed}{HTML}{E9002D}
\definecolor{myAmber}{HTML}{FFAA00}
\definecolor{myGreen}{HTML}{00B000}
\definecolor{myPurple}{HTML}{AF58BA}
\definecolor{myBlue}{HTML}{009ADE}
\definecolor{myGray}{HTML}{A0B1BA}
\definecolor{non-water}{HTML}{5387de}
\definecolor{water}{HTML}{ecb732}
\definecolor{landscape}{HTML}{7d55b2}
\definecolor{foliage}{HTML}{48995f}
\definecolor{gravel}{HTML}{e87b9f}
\definecolor{sky}{HTML}{5cc6db}
\definecolor{cliff}{HTML}{e57439}

\title{\LARGE \bf
Unsupervised RGB-to-Thermal Domain Adaptation via Multi-Domain Attention Network
}

\author{Lu Gan, Connor Lee, and Soon-Jo Chung%
\thanks{*This work is funded by Ford Motor Company and in part by the Office of Naval Research. The authors are with the Division of Engineering and Applied Science,
        California Institute of Technology, Pasadena, CA 91125, USA
        {\tt\small \{ganlu, clee, sjchung\}@caltech.edu}}%
}

\begin{document}

\maketitle
\thispagestyle{empty}
\pagestyle{empty}

\begin{abstract}
This work presents a new method for unsupervised thermal image classification and semantic segmentation by transferring knowledge from the RGB domain using a multi-domain attention network. Our method does not require any thermal annotations or co-registered RGB-thermal pairs, enabling robots to perform visual tasks at night and in adverse weather conditions without incurring additional costs of data labeling and registration. Current unsupervised domain adaptation methods look to align global images or features across domains. However, when the domain shift is significantly larger for cross-modal data, not all features can be transferred. We solve this problem by using a shared backbone network that promotes generalization, and domain-specific attention that reduces negative transfer by attending to domain-invariant and easily-transferable features. Our approach outperforms the state-of-the-art RGB-to-thermal adaptation method in classification benchmarks, and is successfully applied to thermal river scene segmentation using only synthetic RGB images. Our code is made publicly available at ~\href{https://github.com/ganlumomo/thermal-uda-attention}{https://github.com/ganlumomo/thermal-uda-attention}.

\end{abstract}

\section{INTRODUCTION}

Cameras are critical for robot perception as they provide dense measurements and rich environmental information. However, most existing vision models are developed for cameras operating in the visible spectrum due to their ubiquity and the accessibility of large-scale RGB datasets~\cite{deng2009imagenet, lin2014microsoft}. Although these models allow robotic systems such as autonomous vehicles (AV) to work well in ideal conditions with sufficient illumination, their performance is largely degraded at night and in adverse conditions. Thermal cameras, on the other hand, detect electromagnetic waves beyond the visible spectrum that penetrate through dust, smoke, and light fog, enabling around-the-clock robotic operations.

\begin{figure}[t]
    \centering
    \begin{subfigure}{0.2\linewidth}
    \includegraphics[width=1\linewidth]{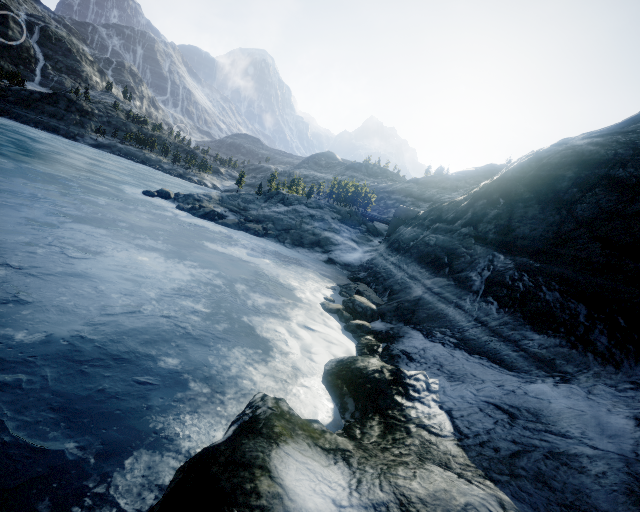}
    \end{subfigure} \hspace{-0.25cm}
    \begin{subfigure}{0.2\linewidth}
    \includegraphics[width=1\columnwidth]{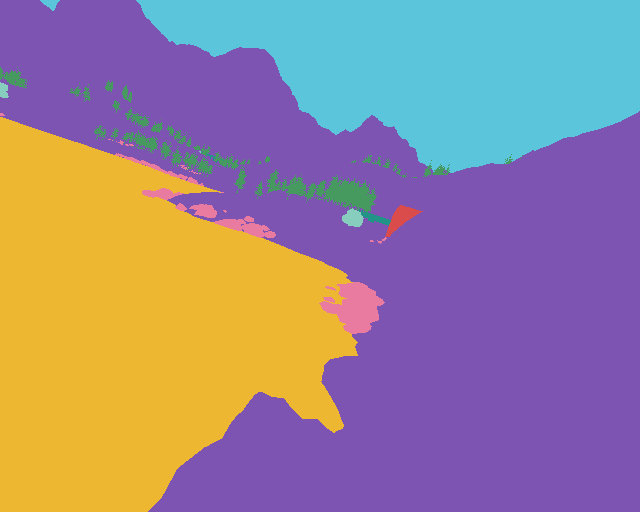}
    \end{subfigure} \hspace{-0.25cm}
    \begin{subfigure}{0.2\linewidth}
    \includegraphics[width=1\columnwidth]{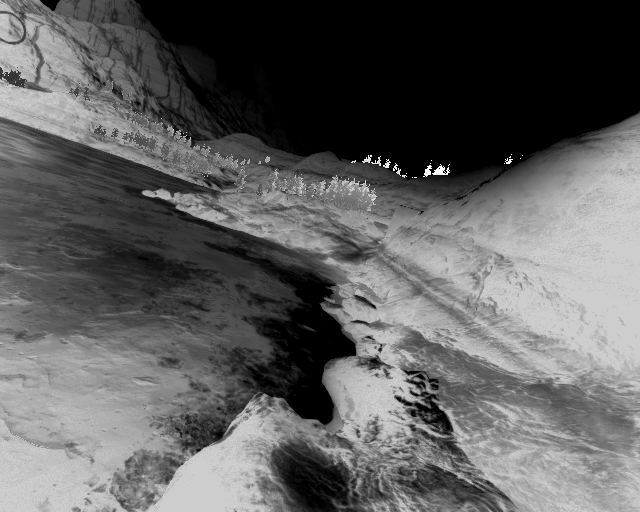}
    \end{subfigure} \hspace{-0.25cm}
    \begin{subfigure}{0.2\linewidth}
    \includegraphics[width=1\columnwidth]{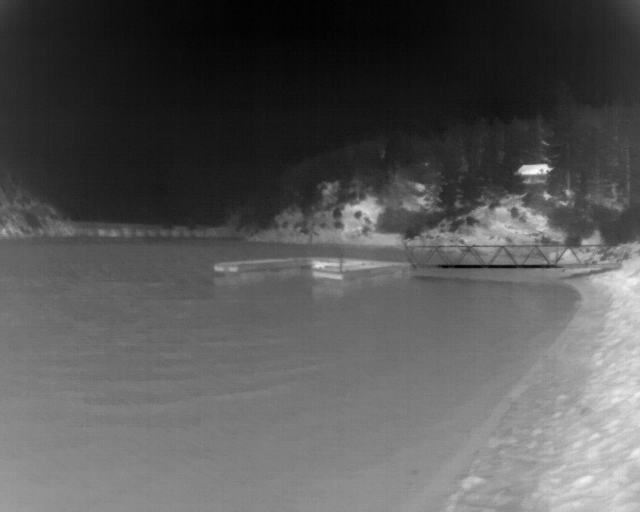}
    \end{subfigure} \hspace{-0.25cm}
    \begin{subfigure}{0.2\linewidth}
    \includegraphics[width=1\columnwidth]{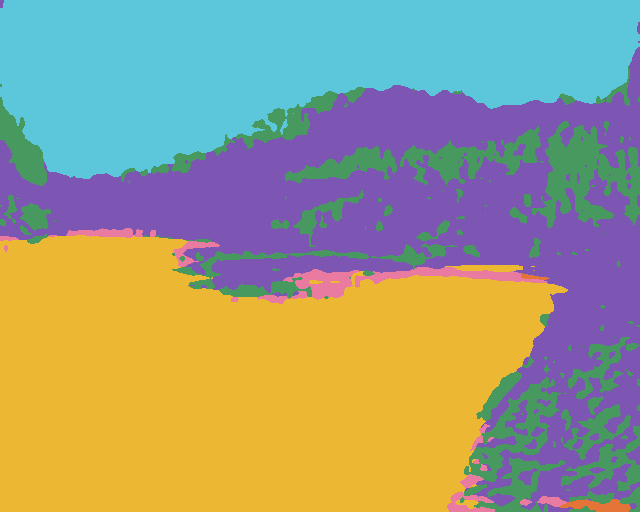}
    \end{subfigure}
    \\
    \begin{subfigure}{0.2\linewidth}
    \includegraphics[width=1\columnwidth]{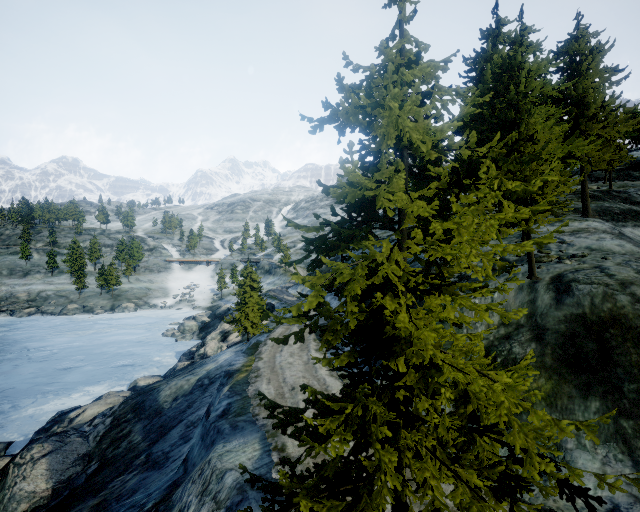}
    \end{subfigure} \hspace{-0.25cm}
    \begin{subfigure}{0.2\linewidth}
    \includegraphics[width=1\columnwidth]{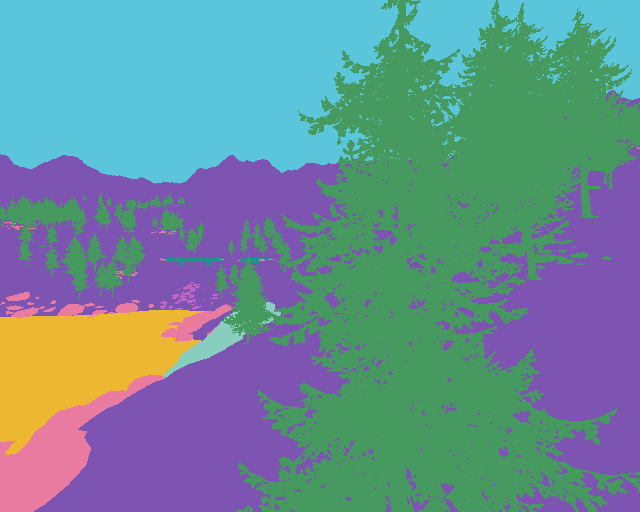}
    \end{subfigure} \hspace{-0.25cm}
    \begin{subfigure}{0.2\linewidth}
    \includegraphics[width=1\columnwidth]{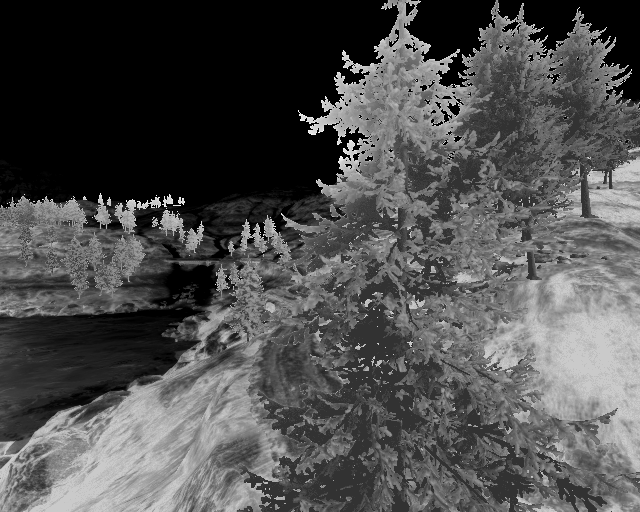}
    \end{subfigure} \hspace{-0.25cm}
    \begin{subfigure}{0.2\linewidth}
    \includegraphics[width=1\columnwidth]{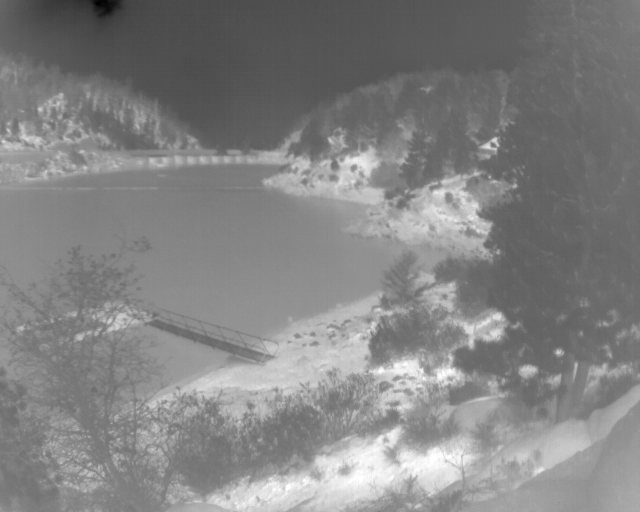}
    \end{subfigure} \hspace{-0.25cm}
    \begin{subfigure}{0.2\linewidth}
    \includegraphics[width=1\columnwidth]{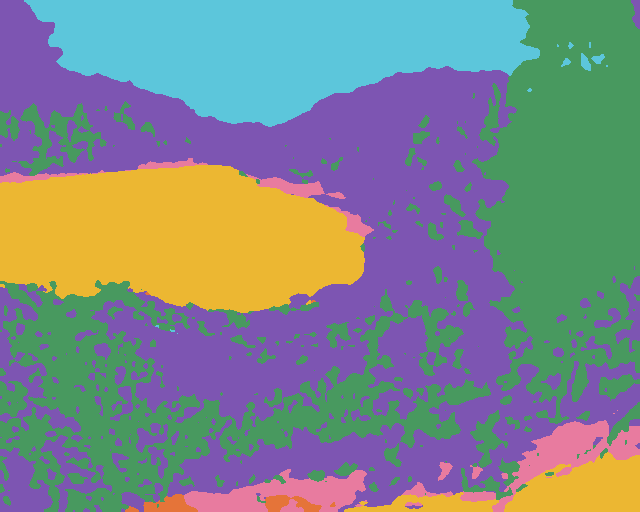}
    \end{subfigure}
    \\
    \begin{subfigure}{0.2\linewidth}
    \includegraphics[width=1\columnwidth]{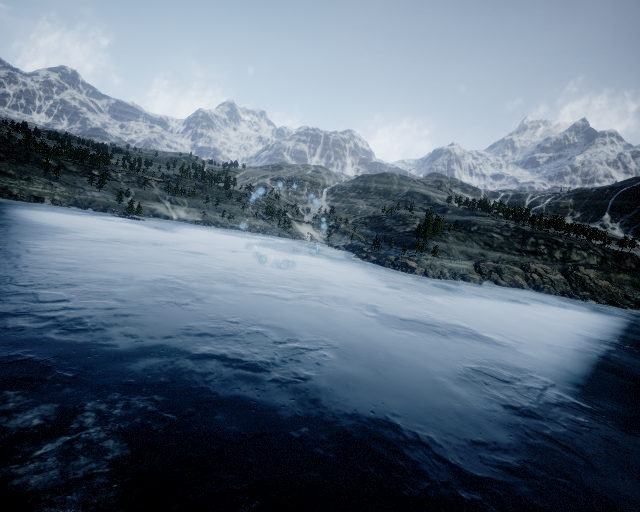}
    \end{subfigure} \hspace{-0.25cm}
    \begin{subfigure}{0.2\linewidth}
    \includegraphics[width=1\columnwidth]{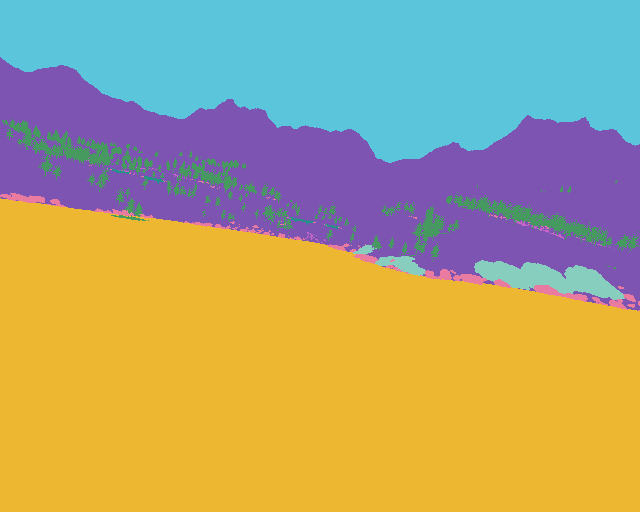}
    \end{subfigure} \hspace{-0.25cm}
    \begin{subfigure}{0.2\linewidth}
    \includegraphics[width=1\columnwidth]{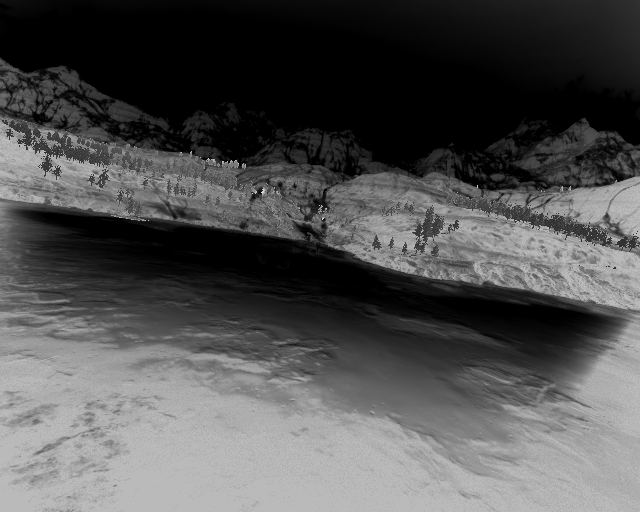}
    \end{subfigure} \hspace{-0.25cm}
    \begin{subfigure}{0.2\linewidth}
    \includegraphics[width=1\columnwidth]{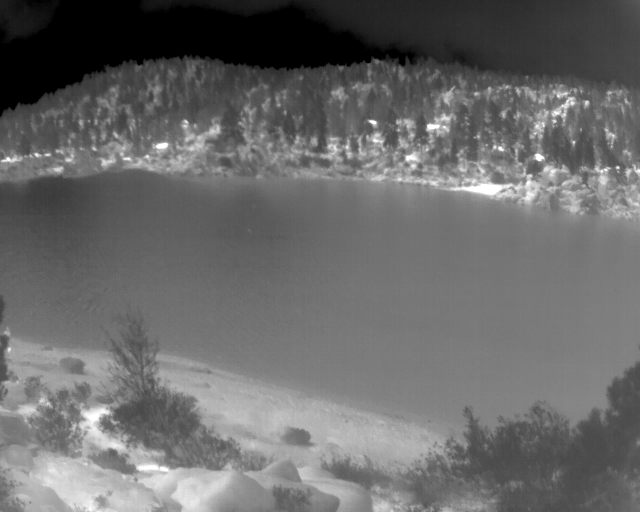}
    \end{subfigure} \hspace{-0.25cm}
    \begin{subfigure}{0.2\linewidth}
    \includegraphics[width=1\columnwidth]{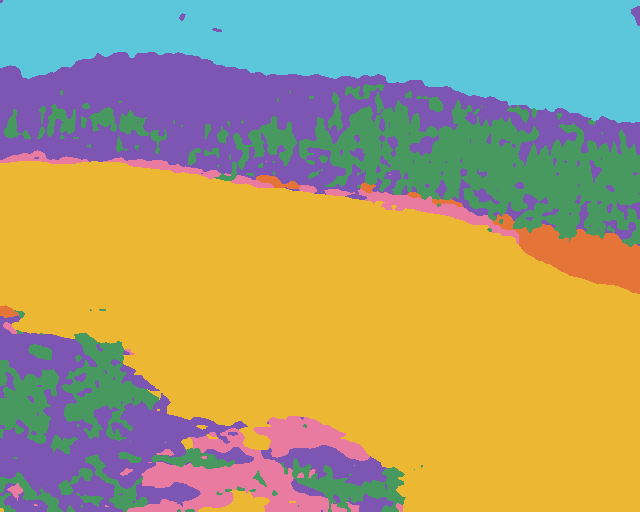}
    \end{subfigure} \\
    \begin{subfigure}{1.00\linewidth} \hspace{-0.165cm}
    \includegraphics[width=\columnwidth, trim={1.7cm, 13.8cm, 0.9cm, 3.8cm}, clip]{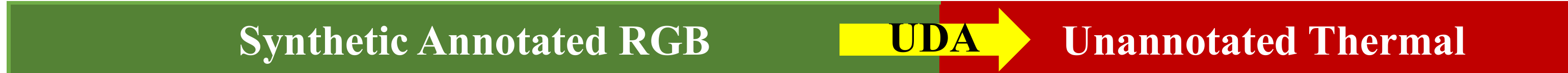}
    \end{subfigure}
    \caption{Our RGB-to-thermal unsupervised domain adaptation (UDA) leverages knowledge learned from a synthetic annotated RGB dataset to perform semantic segmentation on thermal river scenes without requiring thermal annotations.}
    \label{fig:bigbear}
\end{figure}

One popular approach towards robust vision is to leverage thermal images in conjunction with RGB via multi-spectral sensor fusion.
These methods have largely benefited from recent interests in AV technology, resulting in curated datasets~\cite{flir, liu2022target} being made publicly available. Notable examples are GAFF~\cite{zhang2021guided} and CFT~\cite{qingyun2021cross}, two multi-spectral object detection networks trained on paired RGB-thermal image datasets for feature extraction and fusion. In particular, the fusion network in~\cite{qingyun2021cross} sees a $25\%$ performance improvement over a single RGB branch on the FLIR-aligned dataset~\cite{flir}. Urban semantic segmentation has also been improved for nighttime and adverse weather after integrating thermal capabilities~\cite{ha2017mfnet, zhou2022edge, kim2021ms, vertens2020heatnet}. However, these models are fully-supervised, using annotated images or co-registered RGB-thermal pairs which are expensive to acquire and small in scale~\cite{kutuk2022semantic}. In non-AV applications, the lack of thermal data and cost of labeling hinder the development of thermal vision models, especially when current vision models, like Transfomers, have been trending larger~\cite{han2022survey}.

To overcome this issue, we look to leverage existing large-scale RGB datasets to learn thermal models via unsupervised domain adaptation (UDA) techniques. UDA aims to transfer the knowledge learned in a labeled source domain to an unlabeled target domain~\cite{wilson2020survey}. Although most UDA methods focus on domains from different environments but within the same modality (mainly RGB images), such as GTAV-to-Cityscapes~\cite{tsai2018learning, hoffman2018cycada}, the underlying assumption that a domain-invariant feature representation exists also applies to cross-modal data, especially for semantic-related tasks.

In this work, we aim to transfer knowledge learned from labeled RGB images to unlabeled thermal images. This is challenging for two reasons: First, cross-modal domains have larger domain shifts and more dissimilar features compared to domains within same modalities. UDA methods that match global images or feature distributions of both domains can hurt generalization and lead to \emph{negative transfer} in which untransferable features are forcefully aligned~\cite{zhao2019learning, wang2019transferable},~\cite{zhang2020transferable}. Second, UDA methods based on generative adversarial networks (GANs) need a large amount of unlabeled target data to be well-trained~\cite{wilson2020survey} which can also be unavailable in the thermal domain.

We surmount these challenges by designing a multi-domain attention network with a shared backbone and domain-specific attention for RGB-to-thermal adaptation. This shared backbone promotes generalization across domains, prevents feature over-alignment, and relaxes the thermal dataset size requirement. For feature alignment, we train the target-specific attention using adversarial learning to attend to and transfer more domain-invariant and transferable features among all shared features to alleviate negative transfer. The main contributions of our work are as follows:

\begin{itemize}
    \item We establish an unsupervised RGB-to-thermal domain adaptation method using a multi-domain attention network and adversarial attention learning.
    \item We evaluate our method on thermal image classification tasks and outperform the state-of-the-art RGB-to-thermal adaptation approach on two benchmarks.
    \item We demonstrate the versatility of our approach, leveraging it to perform thermal river scene segmentation, and to the best of our knowledge, are the first to utilize synthetic RGB data for thermal semantic segmentation.
\end{itemize}

\section{RELATED WORK}

\textbf{Unsupervised Domain Adaptation:}
UDA has been successfully applied to a variety of vision tasks including image classification~\cite{ganin2015unsupervised, tzeng2017adversarial, saito2018maximum, long2018conditional, sgada2021}, semantic segmentation~\cite{hoffman2018cycada, tsai2018learning, gao2022cross} and 2D/3D object detection~\cite{xu2021spg, marnissi2022unsupervised}. Domain alignment is the fundamental principle of UDA, and can be achieved by two main methodologies: domain mapping and domain-invariant feature learning~\cite{wilson2020survey}. Domain mapping can be viewed as pixel-level alignment which maps images from one domain to another via image translation.
For instance, PixelDA~\cite{bousmalis2017unsupervised} and CyCADA~\cite{hoffman2018cycada} map source training data into the target domain using conditional GANs and train the downstream model on the fake target data.
Pixel-level alignment can remove the domain differences in the input space to some extent but such differences are primarily low-level~\cite{wilson2020survey}. 
Other works achieve domain adaptation by domain-invariant feature learning or feature-level alignment. By mapping source and target input data to the same feature distribution, a downstream predictor trained on such domain-invariant features from source can also work well on the target domain.
This is typically done by minimizing a distance defined on distributions~\cite{saito2018maximum}, or by adversarial training via a domain discriminator that attempts to distinguish between source and target features~\cite{ganin2015unsupervised, tzeng2017adversarial, long2018conditional, tsai2018learning, sgada2021}.
Our method is similar to these works and can be viewed as an instance of the general pipeline in~\cite{tzeng2017adversarial} by leveraging multi-domain network and attention mechanisms.

\textbf{RGB-to-Thermal UDA:}
Despite the success of UDA on visible images, adapting models from visible to thermal remains challenging due to their larger domain gap.
Existing RGB-to-thermal adaptation works like MS-UDA~\cite{kim2021ms} and HeatNet~\cite{vertens2020heatnet} distill knowledge from a semantic segmentation network pretrained on RGB datasets to their two-stream network by pseudo-labeling RGB-thermal image pairs. However, as the pseudo-labels are generated for the RGB image in a pair, the main domain gap here is intra-modal, between the pretraining dataset and RGB images in the paired dataset, rather than inter-modal. 

Our work is mostly related to SGADA~\cite{sgada2021} and Marnissi \emph{et al.}~\cite{marnissi2022unsupervised} which aim to transfer knowledge from RGB to thermal without requiring thermal annotations or RGB-thermal pairs. For pedestrian detection, Marnissi \emph{et al.}~\cite{marnissi2022unsupervised} incorporates alignment at difficult levels into Faster R-CNN~\cite{ren2015faster} using adversarial training.
SGADA~\cite{sgada2021} is built upon ADDA~\cite{tzeng2017adversarial} with an additional self-training procedure. For pseudo-labeling, not only the model prediction and confidence are considered, but also the prediction and confidence from the domain discriminator. It achieves the best results on MS-COCO~\cite{lin2014microsoft} to FLIR ADAS~\cite{flir} adaptation benchmark, however, its performance largely depends on the quality of pseudo labels generated by ADDA.

\textbf{Attention Networks:} Attention mechanisms allow models to dynamically attend to certain parts of the input that are more effective for a task, and become important concepts in neural networks. Attention can be grouped into different types, including sequence attention, channel attention~\cite{hu2018squeeze}, and spatial attention~\cite{woo2018cbam}, etc. For domain adaptation, Wang \emph{et al.}~\cite{wang2019transferable} and Zhang \emph{et al.}~\cite{zhang2020transferable} propose transferable attention networks using self-attention mechanisms to highlight transferable features. The spatial attention they employed attend to different regions in a feature map. Instead, we use channel-wise attention~\cite{hu2018squeeze} to attend to different feature maps and use residual adapters~\cite{rebuffi2017learning} to align them, with the intuition that certain types of features are more transferable than others. The transferability difference in feature types (i.e., channels) should be focused on more than in feature regions (i.e., spatial locations) for cross-modal domains. 

\begin{figure*}[t]
    \centering
    \includegraphics[width=1.8\columnwidth, trim={1cm 9cm 0.2cm 0.9cm}, clip]{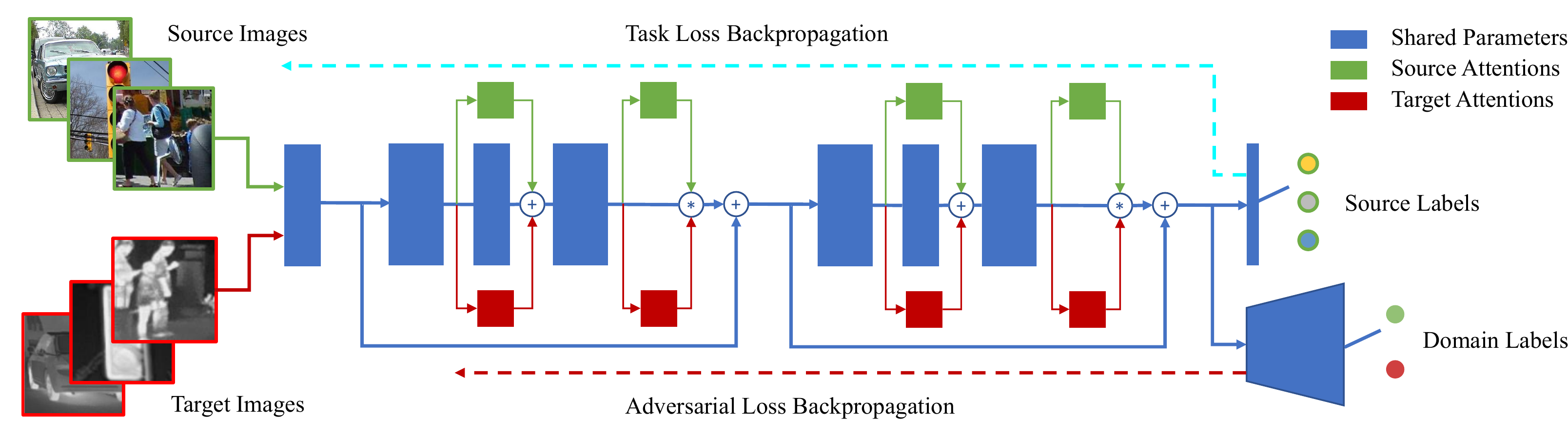}
    \caption{The network architecture and training procedure of our proposed unsupervised RGB-to-thermal domain adaptation method. The specific architecture is shown for image classification task.}
    \label{fig:network}
\end{figure*}

\section{PROPOSED METHOD}

\subsection{Multi-Domain Attention Network}

Our multi-domain attention network design draws ideas from multi-domain learning~\cite{rebuffi2017learning} and task attention mechanisms in multi-task learning~\cite{maninis2019attentive}. Both works use a shared backbone network and domain/task-specific parameters to separate a shared representation learned from all domain/tasks and domain/task-specific modeling capabilities. It has been shown that sharing weights across domains/tasks promotes the generalization ability.
In contrast with encouraging disentanglement in a supervised setup~\cite{rebuffi2017learning, maninis2019attentive}, we use domain-specific attention with adversarial learning to facilitate domain-invariant feature extraction and alignment for domain adaptation. 

Our multi-domain attention network consists of an encoder-decoder backbone, shared by both source and target domains, with domain-specific attention modules attached at various stages of the encoder. For UDA classification (Fig.~\ref{fig:network}), the architecture consists of the shared backbone and classifier (blue), source-specific (green), and target-specific (red) attention modules. Hypothesizing that different sensor modality favors different types of features, we use channel-wise attention, i.e., Squeeze-and-Excitation (SE)~\cite{hu2018squeeze}, to highlight more domain-invariant and easily-transferable feature maps among all shared features, and use residual adapters~\cite{rebuffi2017learning} to align them across domains.

Let \mbox{$F_c \in \mathbb{R}^{h \times w \times C^{\prime}}$} denote a convolutional layer of $C$ kernels of size $h \times w$ operating on $C^{\prime}$ input channels, we have \mbox{$F_c : x \rightarrow f$}, \mbox{$x \in \mathbb{R}^{H^{\prime} \times W^{\prime} \times C^{\prime}}$}, \mbox{$f \in \mathbb{R}^{H \times W \times C}$}, where \mbox{$f = [f_1, f_2, ..., f_C]$} represent $C$ output feature maps. A SE module~\cite{hu2018squeeze} first ``squeezes'' $f$ into a low-dimensional channel descriptor \mbox{$d \in \mathbb{R}^{{\frac{C}{r}}}$} with reduction ratio $r$. This is done using global average pooling followed by a fully connected (FC) layer with ReLU activations. The channel descriptor is then transformed into channel-wise weight coefficients \mbox{$s = [s_1, s_2, ... s_C], s_c \in (0,1)$} through another FC layer and a sigmoid function. Finally, $s$ is used to ``excite'' different feature maps in $f$ by feature channel reweighting: $\widetilde{f}_c = s_c \cdot f_c$. In our network, we use domain-specific SE blocks operating on the shared feature maps right before the residual addition in residual blocks, as shown in Fig.~\ref{fig:network}.

By attaching domain-specific SE modules to the shared backbone network, they have the ability to accentuate more domain-invariant and transferable features in the shared features while attenuate less-transferable ones. To further align the reweighted features across domains, we leverage residual adapters~\cite{rebuffi2017learning} to directly and dynamically adapting the shared feature extractors $F_c$ to domain-specific feature extractors $G_d$. Specifically, $d=c$ in our network.

Given a shared convolutional layer of $C$ kernels \mbox{$F_c \in \mathbb{R}^{h \times w \times C^{\prime}}$}, a domain-specific convolutional layer with $D$ filters \mbox{$G_d \in \mathbb{R}^{h \times w \times C^{\prime}}$} can be simply constructed as an affine transformation of $F_c$ using only a small amount of additional parameters \mbox{$\alpha = \{ \alpha_{dc} \}$}:
\begin{equation}
    G_d = \sum_{c=1}^C \alpha_{dc} F_c.
\end{equation}
Here, \mbox{$\alpha \in \mathbb{R}^{D \times C}$} are the trainable residual adapter parameters~\cite{rebuffi2017learning}. This linear parameterization reduces constructing $G_d$ for each domain to the shared $F_c$ with a small amount of domain-specific parameters $\alpha$. The works~\cite{rebuffi2017learning,rebuffi2018efficient} further show that $\alpha$ can be reparameterized and implemented as a convolutional layer of $1 \times 1$ filters connected in parallel with the shared convolutional layer. In our network, we add residual adapters to the middle $3 \times 3$ convolutional layer in residual blocks for feature alignment, as shown in Fig.~\ref{fig:network}.

We emphasize the differences of our attention modules from those in~\cite{rebuffi2017learning, maninis2019attentive}. The multi-domain learning in~\cite{rebuffi2017learning} and multi-task learning in~\cite{maninis2019attentive} are essentially supervised. Their objective is to learn a domain/task-invariant feature representation $f_{inv}$ and domain/task-specific attention $\theta_a, \theta_b$, so that \mbox{$\theta_a(f_{inv}) = f_a$}, \mbox{$\theta_b(f_{inv}) = f_b$}, where $f_a$ and $f_b$ are features tailored for domain/task A and B respectively. In contrast, for our UDA problem, we learn discriminative features $f_s$ for a given task using supervised training in the source domain and target-specific attention $\theta_t$ using adversarial training for feature alignment, i.e. \mbox{$\theta_s(f_{sh}) = f_s$}, \mbox{$\theta_t(f_{sh}) = f_{t \rightarrow s}$}, where $\theta_s$ and $\theta_t$ are source and target attention, $f_{sh}$ and $f_{t \rightarrow s}$ are the shared features and the target features aligned with $f_s$ respectively.

\subsection{Adversarial Attention Learning}
\label{sec:training}
To perform unsupervised domain adaptation, we train subsets of network parameters in an alternating fashion. We denote the shared parameters, including the backbone network and the decoder, as $\theta_{sh}$, and the source- and target-specific attention modules as $\theta_s$ and $\theta_t$ respectively. We train $\theta_{sh}$ and $\theta_s$ using labeled data from the source domain and train the task-specific attention modules $\theta_t$ adversarially in an alternating fashion.

Let $M$ denote the proposed multi-domain attention network, and let \mbox{$\mathcal{D}_s = \{(x_s^i, y_s^i)\}_{i=1}^{n_s}$} and \mbox{$\mathcal{D}_t = \{(x_t^j)_{j=1}^{n_t}\}$} denote the annotated training data in the source domain and the unlabeled target training data respectively. The shared and source-specific network parameters can be trained with supervision by minimizing the standard cross-entropy loss. For a classification problem, the loss can be written as:
\begin{equation}
\label{eq:loss_cls}
    \mathcal{L}_{task}(x_s, y_s) = - \sum_{c=1}^C \mathds{1}_{[c=y_s]} \log{M(x_s; \theta_{sh+s})},
\end{equation}
where $(x_s, y_s)$ are source data-label pairs drawn from $\mathcal{D}_s$, $\mathds{1}_{[x]}$ is an indicator function so that \mbox{$\mathds{1}_{[x]} = 1$} if $x=1$, and 0 otherwise, and $C$ is the number of categories.

\begin{algorithm}[t]
    \caption{Multi-Domain Attention Network for Unsupervised Domain Adaptation}
    \label{alg:train}
    \begin{algorithmic}[1]
    \State \textbf{Input:} \text{Training data:} $\mathcal{D}_s, \mathcal{D}_t$,
    \State \hspace{1cm} \text{Network}: $M = \{\theta_{sh}, \theta_s, \theta_t\}$, \text{Discriminator}: $D$
    \State \hspace{1cm} \text{Learning rate:} $\alpha, \beta, \gamma$
    \State \text{Initialize} $M^0, D^0$
    \For{$n$ = 1 to $N$}
        \State Sample batch data $(x_s, y_s)$ from $\mathcal{D}_s$, and $x_t$ from $\mathcal{D}_t$
        \State $l_{task} \gets \mathcal{L}_{task}(x_s, y_s)$ \Comment{Evaluate~\eqref{eq:loss_cls}}
        \State $\theta_{sh+s}^n = \theta_{sh+s}^{n-1} - \alpha \nabla_{\theta_{sh+s}^{n-1}}l_{task}$
        \State $l_{adv} \gets \mathcal{L}_{adv}(x_t, D^{n-1})$ \Comment{Evaluate~\eqref{eq:loss_adv}}
        \State $\theta_{t}^n = \theta_{t}^{n-1} - \beta \nabla_{\theta_{t}^{n-1}}l_{adv}$
        \State $l_{dis} \gets \mathcal{L}_{dis}(x_s, x_t, M^n)$ \Comment{Evaluate~\eqref{eq:loss_dis}}
        \State $D^n = D^{n-1} - \gamma \nabla_{D^{n-1}}l_{dis}$
    \EndFor
    \State \textbf{Output:} $M^N, D^N$
    \end{algorithmic}
\end{algorithm}

We train the target-specific attention in our network adversarially by forcing the target attention to attend to domain-invariant features from the shared features and further align them with the source feature distributions. Adversarial attention learning can be achieved by approaching the following minimax game~\cite{ganin2016domain, tzeng2017adversarial} between the target-specific attention $\theta_t$ and a domain discriminator $D$:
\begin{align}
\label{eq:adv}
\min_{\theta_t} \max_{D} &\mathcal{L}(D, \theta_t) = \\ \nonumber
& \mathbb{E}_{x_s \sim \mathcal{D}_s} \log D(f_s) + \mathbb{E}_{x_t \sim \mathcal{D}_t} \log (1 - D(f_t)),
\end{align}
where $f_s$ and $f_t$ are source and target features from the entire encoder with weights $\theta_{sh+s}$ and $\theta_{sh+t}$, respectively.

Specifically, the minimax loss in~\eqref{eq:adv} is split into two objectives, where the domain discriminator plays the adversarial role and attempts to distinguish between source features $f_s$ and target features $f_t$ by minimizing the following loss:
\begin{equation}
\label{eq:loss_dis}
    \mathcal{L}_{dis}(x_s, x_t, M) = -\log D(f_s) -\log{(1-D(f_t))},
\end{equation}
and the target-specific attention is trained to fool the domain discriminator and increase domain confusion by minimizing an adversarial loss:
\begin{equation}
\label{eq:loss_adv}
    \mathcal{L}_{adv}(x_t, D) = - \log{D(f_t)}.
\end{equation}
The three-step training procedure of our multi-domain attention network is given in Algorithm~\ref{alg:train}.

Advantages of this alternating training are twofold. First, when training $\theta_{sh+s}$ using $\mathcal{L}_{task}(x_s, y_s)$, it reduces to training a supervised source model and the feature extractor learns to extract the most discriminative features for the given task. When training $\theta_{t}$ with fixed $\theta_{sh}$, $\theta_{t}$ learns to select and adapt the most domain-invariant ones among the discriminative features, leading to better adaptation performance. Second, it eliminates a weighting hyperparameter for two loss functions and makes the training procedure more stable.

\subsection{Self-Training}

We further fine-tune the model with a single self-training step using the pseudo labels generated for the target training data. Following~\cite{sgada2021}, we save the prediction and corresponding confidence (the maximum of softmax probabilities) of the model $M$ trained in Sec.~\ref{sec:training} for all unlabeled target training samples. In the meantime, the prediction and confidence of the domain discriminator $D$ are also recorded. For target samples that successfully fool the discriminator ($D$ predicts them as source samples with a high confidence), we assign them pseudo-labels according to the model prediction. For those target samples that the discriminator recognize but with low domain confidence, we also include them in pseudo-labeling. The pseudo-labels are further filtered by the model prediction confidence.

In this stage, we only train the target-specific attention parameters using a cross-entropy loss in a supervised setup:
\begin{equation}
\label{eq:loss_st}
    \mathcal{L}_{st}(x_t, \hat{y}_t) = - \sum_{c=1}^C \mathds{1}_{[c=\hat{y}_t]} \log{M(x_t; \theta_{t})},
\end{equation}
where $\hat{y}_t$ is the generated pseudo-label for target training data $x_t$. This way, we can further improve the performance in target while keeping the performance in source, so that we have a single unified model that performs well on both source and target data. This learning-without-forgetting~\cite{li2017learning} property is another benefit of our multi-domain attention network.

\section{EXPERIMENTAL RESULTS}

\subsection{Implementation}
\label{sec:implementation}
For a fair comparison, we employ the same backbone architecture used in other methods, i.e. a ResNet-50 pretrained on ImageNet~\cite{deng2009imagenet}. We use a FC classifier and a FC discriminator for image classification, and use an Atrous Spatial Pyramid Pooling~\cite{chen2017deeplab} decoder and a fully-convolutional discriminator for semantic segmentation. Parameters are all updated using the ADAM optimizer with \mbox{$\beta_1 = 0.5, \beta_2=0.999$} and weight decay of $2.5 \times 10^{-5}$. The learning rate $\alpha, \beta, \gamma$ in Algorithm~\ref{alg:train} is set to as $1 \times 10^{-4}, 1 \times 10^{-5}$ and $1\times10^{-3}$, respectively. All experiments are conducted on a NVIDIA Quadro RTX 8000 GPU with 48GB memory.

\begin{figure}[t]
    \centering
    \includegraphics[width=0.15\columnwidth]{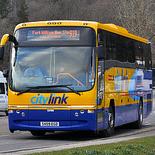}
    \includegraphics[width=0.15\columnwidth]{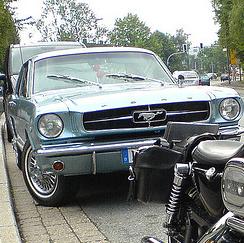}
    \includegraphics[width=0.15\columnwidth]{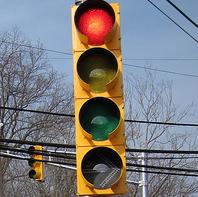}
    \includegraphics[width=0.15\columnwidth]{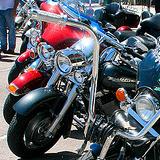}
    \includegraphics[width=0.15\columnwidth]{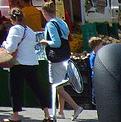}
    \includegraphics[width=0.15\columnwidth]{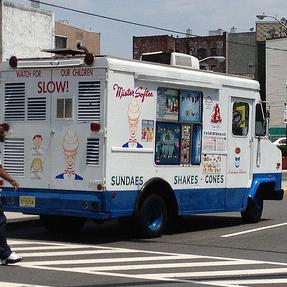} \\
    \includegraphics[width=0.15\columnwidth]{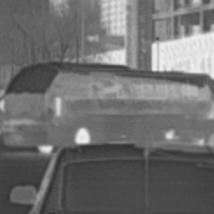}
    \includegraphics[width=0.15\columnwidth]{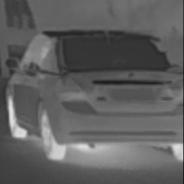}
    \includegraphics[width=0.15\columnwidth]{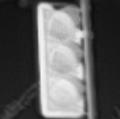}
    \includegraphics[width=0.15\columnwidth]{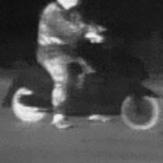}
    \includegraphics[width=0.15\columnwidth]{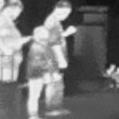}
    \includegraphics[width=0.15\columnwidth]{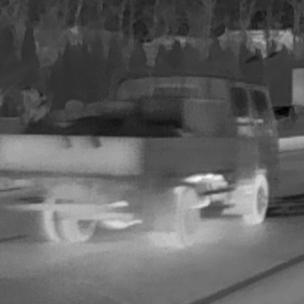}
    \caption{Examples of the prepared data from MS-COCO~\cite{lin2014microsoft} and $\text{M}^3$FD Detection datasets~\cite{liu2022target}.}
    \label{fig:m3fd}
\end{figure}

\begin{table}[t]
\caption{Data statistics of MS-COCO to $\text{M}^3$FD dataset.}
\label{tab:m3fd_data}
    \centering
    \resizebox{\columnwidth}{!}{ 
    \begin{tabular}{cc|ccccccc}
    \hline
    & & \rot{Bus} & \rot{Car} & \rot{Light} & \rot{Motor.} & \rot{People} & \rot{Truck} & \rot{Total}\\
    \hline
    \multirow{2}{*}{MS-COCO} & Train & 3887 & 36830 & 12139 & 6330 & 200831 & 7232 & 267249 \\
    & Val & 189 & 1623 & 603 & 229 & 8331 & 315 & 11290 \\
    \hline
    \multirow{3}{*}{$\text{M}^{3}$FD} & Train & 441 & 12969 & 1902 & 382 & 8770 & 696 & 25160 \\
    & Val & 55 & 1621 & 238 & 48 & 1096 & 87 & 3141 \\
    & Test & 55 & 1620 & 237 & 47 & 1096 & 86 & 3141 \\
    \hline
    \end{tabular}
    }
\end{table}

\begin{figure*}[t]
\centering
\begin{tikzpicture}
\matrix (a)[row sep=0mm, column sep=0mm, inner sep=0mm,  matrix of nodes] at (0,0) {
        \includegraphics[width=0.16\columnwidth]{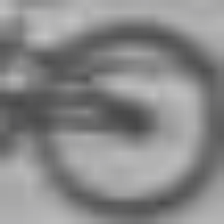} &
        \includegraphics[width=0.16\columnwidth]{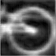} &
        \includegraphics[width=0.16\columnwidth]{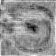} &
        \includegraphics[width=0.16\columnwidth]{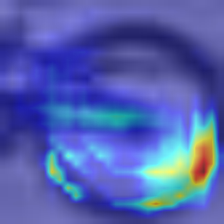} \hspace{0.5mm} & 
        \hspace{0.5mm} \includegraphics[width=0.16\columnwidth]{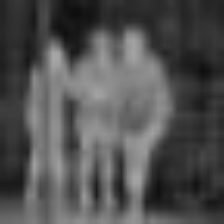} &
        \includegraphics[width=0.16\columnwidth]{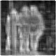} &
        \includegraphics[width=0.16\columnwidth]{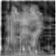} &
        \includegraphics[width=0.16\columnwidth]{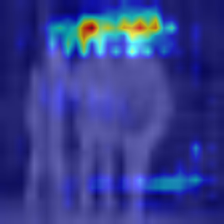} \hspace{0.5mm} &
        \hspace{0.5mm}
        \includegraphics[width=0.16\columnwidth]{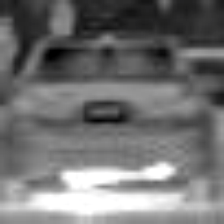} &
        \includegraphics[width=0.16\columnwidth]{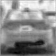} &
        \includegraphics[width=0.16\columnwidth]{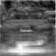} &
        \includegraphics[width=0.16\columnwidth]{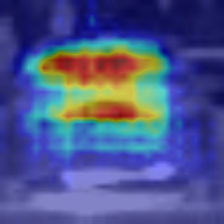} \\
        \includegraphics[width=0.16\columnwidth]{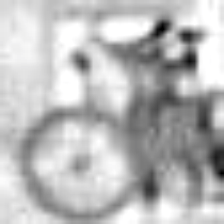} &
        \includegraphics[width=0.16\columnwidth]{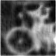} &
        \includegraphics[width=0.16\columnwidth]{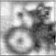} &
        \includegraphics[width=0.16\columnwidth]{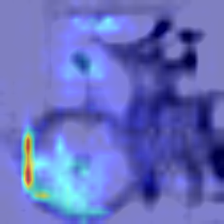} \hspace{0.5mm} &
        \hspace{0.5mm} \includegraphics[width=0.16\columnwidth]{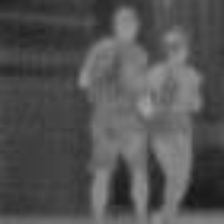} &
        \includegraphics[width=0.16\columnwidth]{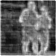} &
        \includegraphics[width=0.16\columnwidth]{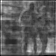} &
        \includegraphics[width=0.16\columnwidth]{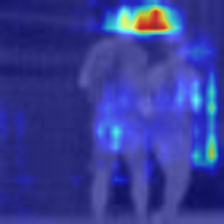} \hspace{0.5mm} &
        \hspace{0.5mm} \includegraphics[width=0.16\columnwidth]{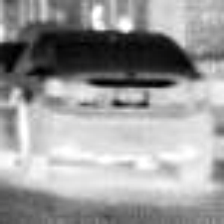} &
        \includegraphics[width=0.16\columnwidth]{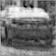} &
        \includegraphics[width=0.16\columnwidth]{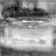} & \includegraphics[width=0.16\columnwidth]{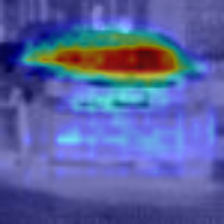} \\
    };
\draw[thick,densely dashed, blue] (a-1-4.north east) -- (a-2-4.south east);
\draw[thick,densely dashed, blue] (a-1-8.north east) -- (a-2-8.south east);
\end{tikzpicture}
\caption{Task-specific attention visualization. From left to right in each block: input image, feature map with highest SE weight, feature map with the lowest SE weight, and Grad-CAM~\cite{selvaraju2017grad} of residual adapters for that class.}
\label{fig:attention}
\end{figure*}

\begin{table}[t]
\caption{Ablation study of different attention modules and training strategies on MS-COCO to FLIR ADAS dataset.}
\label{tab:ablation}
\begin{center}
\resizebox{0.85\columnwidth}{!}{
\begin{tabular}{ccc|cccc}
\hline
& Adapter & SE & Bicycle & Car & Person & Average \\
\hline
\multirow{3}{*}{$\theta_{sh+s+t}$} & \checkmark & & 89.43 & \bf 97.14 & 88.89 & \bf 91.83 \\
& & \checkmark & \bf 91.72 & 93.79 & 83.96 & 89.83 \\
& \checkmark & \checkmark & 87.82 & 94.34 & \bf 91.52 & 91.23 \\
\hline
\multirow{3}{*}{$\theta_{sh}$, $\theta_{s+t}$} & \checkmark & & 81.84 & 96.36 & 96.66 & \bf 91.62 \\
& & \checkmark & \bf 82.03 & 93.18 & 95.84 & 90.35 \\
& \checkmark & \checkmark & 79.54 & \bf 96.69 & \bf 96.99 & 91.07 \\
\hline
\multirow{3}{*}{$\theta_{sh+s}$, $\theta_{t}$} & \checkmark & & \bf 90.57 & 97.22 & 89.83 & 92.54 \\ 
 & & \checkmark & 85.75 & 97.48 & \bf 95.85 & 93.03 \\
& \checkmark & \checkmark & 89.20 & 96.87 & 95.59 & \bf 93.88 \\
\hline
\end{tabular}
}
\end{center}
\end{table}

\subsection{Ablation Study}

To investigate the effects of different attention modules and training strategies on adaptation performance, we conduct a thorough ablation study on 9 training combinations resulting from two types of attention modules, i.e. with ($\checkmark$) and without the residual adapter/SE module, and 3 different strategies to train them:
\begin{enumerate}
    \item $\theta_{sh+s+t}$: We jointly train all network parameters using the sum of $\mathcal{L}_{task}$ in~\eqref{eq:loss_cls} and $\mathcal{L}_{adv}$ in~\eqref{eq:loss_adv}, reducing the three training steps in Algorithm~\ref{alg:train} to only alternatively training the model $M$ and domain discriminator $D$.

    \item $\theta_{sh}, \theta_{s+t}$: We alternatively train the shared parameters $\theta_{sh}$ and all domain-specific parameters~$\theta_{s+t}$. In this setting, only $\theta_{sh}$ is updated using $\mathcal{L}_{task}$, while $\theta_{s+t}$ are adversarially trained using a cross-entropy domain loss in~\cite{tzeng2015simultaneous} instead of $\mathcal{L}_{adv}$ in Algorithm~\ref{alg:train}.
    \item $\theta_{sh+s}, \theta_{t}$: The training procedure given in Algorithm~\ref{alg:train}.
\end{enumerate}

Table~\ref{tab:ablation} lists the ablation study results using top-1 accuracy. From the table, alternatively training $\theta_{sh+s}$ and $\theta_{t}$ has better performance compared with the other two training strategies, and with both residual adapter and SE, it achieves the best result among all configurations. This observation aligns with the discussion in Sec.~\ref{sec:training}. In the following experiments, we use setting 3 with both attention modules for our method.

\begin{figure}[t]
    \centering
    \includegraphics[width=0.95\columnwidth]{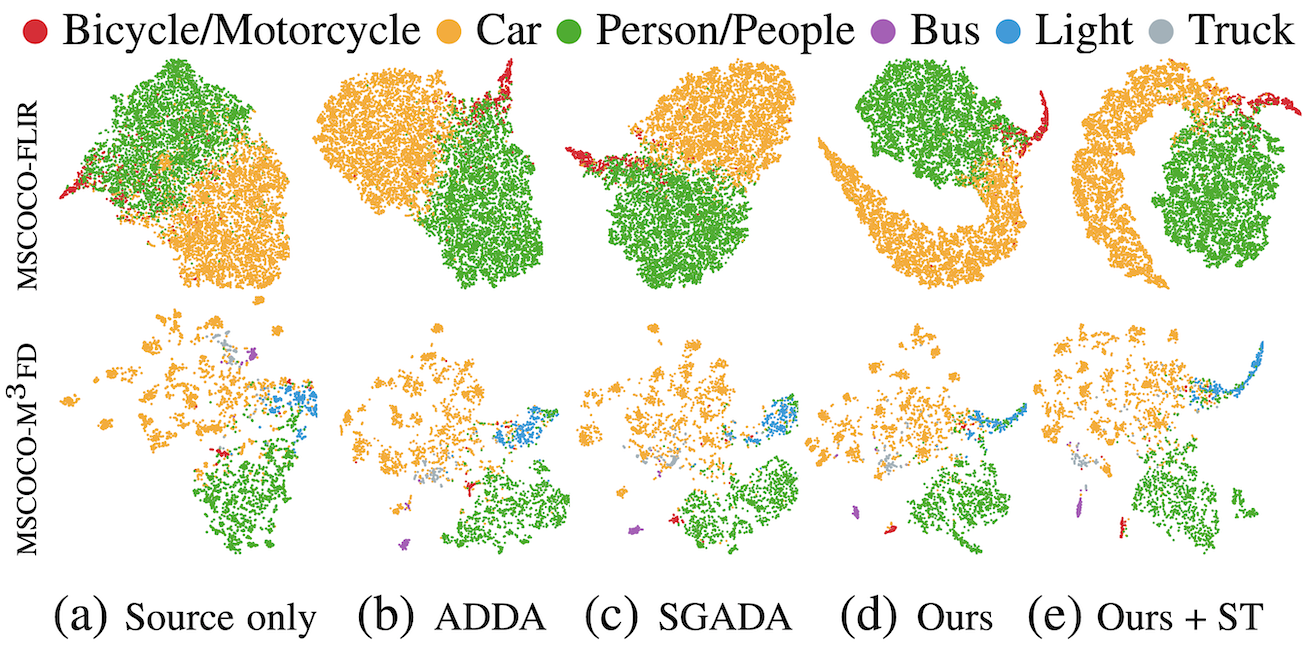}
    \caption{The t-SNE visualization of the encoded features of all target test samples by different methods (ST: Self-training).}
    \label{fig:tsne}
\end{figure}

\subsection{Unsupervised Thermal Image Classification}

\textbf{MS-COCO to FLIR ADAS:} We first compare our method with SGADA~\cite{sgada2021} which achieves the best performance on MS-COCO to FLIR ADAS classification benchmark~\cite{mscoco-flir} and several other the state-of-the-art general UDA methods. MS-COCO~\cite{lin2014microsoft} is a large-scale RGB dataset and FLIR ADAS~\cite{flir} is a popular thermal image dataset for urban environments. We use the dataset prepared by~\cite{sgada2021} including three categories, i.e., bicycle, car and person, in this experiment. Same as~\cite{sgada2021}, we train our network for 15 epochs with a batch size of 32. Per-class accuracy of all methods are given in Table~\ref{tab:flir}, where the proposed method outperforms other methods by a significant margin even without self-training.

\textbf{MS-COCO to $\textbf{M}^3$FD:} MS-COCO to FLIR ADAS dataset has 633440 unannotated target samples~\cite{sgada2021}. To further evaluate the adaptation performance when target training samples are scarce, we prepare a new RGB-to-thermal adaptation benchmark using MS-COCO and $\text{M}^3$FD~\cite{liu2022target} including 6 categories for evaluation, following the data preparation process in~\cite{sgada2021}. Examples and statistics of the prepared dataset are given in Fig.~\ref{fig:m3fd} and Table~\ref{tab:m3fd_data}. Due to less training data, we train all networks for 30 epochs using a batch size of 32. We have similar observations from Table~\ref{tab:m3fd} as from previous experiment, except that all methods outperform the target only model which shows the effectiveness of UDA when sufficient annotated data is unavailable.

\textbf{Experiment Analysis:} We visualize the feature representations for all test samples on target domain using t-SNE~\cite{van2008visualizing} in Fig.~\ref{fig:tsne}, where the better feature separation in (d) and (e) shows our method can learn discriminative features for the given task. To examine the effectiveness of attention modules in our method, we further visualize the trained target-specific attentions in Fig.~\ref{fig:attention} by plotting the features they attend to. For SE modules, we plot the feature map with the highest and the lowest attention weight in the first residual block. As for residual adapters, we plot the Grad-CAM~\cite{selvaraju2017grad} of the last task-specific adapter layer.

We have several interesting observations. First, for bicycle and person categories, the feature maps that the SE highlights the most tend to have high activation on the object contour, which suggests that contour-sensitive features are more domain-invariant and transferable between RGB and thermal domains, aligning with the conclusion in~\cite{chen2022light}. Second, we notice that cars in thermal images are usually brighter at the bottom due to the high temperature in those regions, as opposed to cars in RGB images which appear darker at the bottom due to shadows. The feature maps that the SE module attends to eliminate this phenomenon and appear visually similar to that of cars from an RGB image. Those observations show the effectiveness of our attention network in extracting domain-invariant and transferable features.

\begin{table}[t]
\caption{Top-1 accuracy for MS-COCO to FLIR ADAS.}
\label{tab:flir}
\begin{center}
\resizebox{0.9\columnwidth}{!}{
\begin{tabular}{c|cccc}
\hline
Method & Bicycle & Car & Person & Average \\
\hline
Source only & 69.89 & 83.89 & 86.52 & 80.10 \\
Pixel-DA~\cite{bousmalis2017unsupervised} & 62.53 & 89.99 & 76.73 & 76.42 \\
DTA~\cite{lee2019drop} & 75.45 & \bf 97.65 & 92.45 & 88.52 \\
MCD-DA~\cite{saito2018maximum} & 81.71 & 94.90 & 91.83 & 89.48 \\
DANN~\cite{ganin2015unsupervised} & 78.16 & 95.07 & 96.24 & 89.82 \\
CDAN~\cite{long2018conditional} & 78.16 & 97.10 & 94.82 & 90.03 \\
ADDA~\cite{tzeng2017adversarial} & 86.67 & 96.95 & 89.10 & 90.90 \\
SGADA~\cite{sgada2021} & 87.13 & 94.44 & 92.03 & 91.20 \\
Ours & 89.20 & 96.87 & 95.59 & 93.88 \\
Ours + ST & \bf 89.63 & 97.06 & \bf 96.03 & \bf 94.24 \\
\hline
Target only & 87.59 & 98.78 & 96.35 & 94.24 \\
\hline
\end{tabular}
}
\end{center}
\end{table}

\begin{table}[t]
    \caption{Top-1 accuracy for MS-COCO to $\text{M}^{3}$FD.}
    \label{tab:m3fd}
    \begin{center}
        \resizebox{\columnwidth}{!}{
        \begin{tabular}{c|cccccc|c}
            \hline
            Method &\rotatebox{40}{Bus} & \rotatebox{30}{Car} & \rotatebox{30}{Light} & \rotatebox{30}{Motor.} & \rotatebox{30}{People} & \rotatebox{30}{Truck} & \rotatebox{30}{Average} \\
            \hline
            Source only & 63.64 & 76.98 & 91.14 & 4.26 & 94.07 & 56.98 & 64.51 \\
            MCD-DA~\cite{saito2018maximum} & 89.09 & 76.98 & \bf 95.36 & \bf 76.59 & 93.89 & 30.23 & 77.00 \\
            DANN~\cite{ganin2015unsupervised} & 89.09 & 82.72 & 51.90 & 68.09 & 92.15 & 74.42 & 76.4 \\
            CDAN~\cite{long2018conditional} & 89.09 & \bf 88.58 & 72.15 & 46.81 & 93.98 & 45.35 & 72.7 \\
            ADDA~\cite{tzeng2017adversarial} & \bf 96.36 & 85.86 & 60.34 & 51.06 & 76.73 & \bf 87.21 & 76.26 \\
            SGADA~\cite{sgada2021} & 94.55 & 87.22 & 70.04 & 51.06 & 77.01 & 81.40 & 76.88 \\
            Ours & 90.91 & 85.37 & 72.57 & 74.47 & 93.80 & 51.16 & 78.05 \\
            Ours + ST & 90.91 & 84.26 & 85.65 & 70.21 & \bf 95.44 & 56.98 & \bf 80.57\\
            \hline
            Target only & 94.55 & 92.53 & 83.12 & 21.28 & 90.24 & 20.93 & 67.11 \\
            \hline
        \end{tabular}
        }
    \end{center}
\end{table}

\subsection{Unsupervised Thermal River Scene Segmentation}

We present an effective and inexpensive approach for thermal semantic segmentation by adapting from synthetic RGB images using the proposed method, and test on thermal river scene segmentation. We collect 8 sequences of thermal images at 60~Hz using a hand-held FLIR ADK Longwave Infrared (LWIR) thermal camera with an NUC Ruby Mini PC at Big Bear Lake, CA. We sample images every 100 frames from the collected 48676 sequential frames and form an unlabeled training set of 486 thermal images. As our ultimate goal is to enable the nighttime coastline exploration ability of our aerial robots~\cite{yang2017vision, meier2021river} by thermal river segmentation, we manually annotate 282 diverse test images with pixel-level ground truth water labels for evaluation. Examples of collected thermal images are shown in Fig.~\ref{fig:bigbear} (4th column).

Due to the lack of annotated RGB dataset for natural scenes similar to our riverine environment, we generate synthetic RGB images with automatically obtained semantic labels using the AirSim simulator~\cite{airsim2017fsr}. To that end, we use a publicly available simulation environment, i.e. the Landscape Mountains, and simulate a drone platform with a mounted RGB camera to follow a simple survey trajectory around rivers using the built-in simple flight controller. Following our thermal camera, we set the simulated RGB camera to have 75-degree FoV and capture $640 \times 480$ images. We acquire RGB images and the corresponding semantic labels at 2Hz, and obtain a synthetic labeled river scene RGB dataset of 1357 samples. We further convert the RGB images to grayscale and invert the intensity values (except for the foliage class), resulting in training samples visually close to our thermal images. Examples of the synthetic RGB images, semantic labels, and inverted grayscale images are shown in the first three columns of Fig~\ref{fig:bigbear}.

\begin{figure}[t]
    \centering
    \includegraphics[width=0.95\columnwidth]{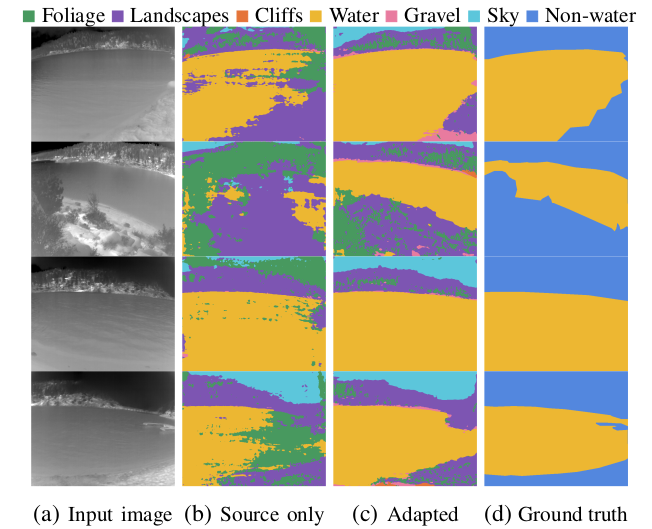}
    \caption{Qualitative results of our unsupervised thermal river segmentation model adapted from synthetic RGB data.}
    \label{fig:seg}
\end{figure}

We train the network for 50 epochs using a batch size of 8 without performing self-training. From Table~\ref{tab:water_seg}, our adapted model obtains a performance gain of $14.6\%$ mIoU and $21.87\%$ water-class IoU over the source only model. The effectiveness of our method can be also seen in Fig.~\ref{fig:bigbear} and Fig.~\ref{fig:seg}, where the adapted model corrects a large portion of false positive foliage prediction. This experiment demonstrates that thermal vision models can be effectively learned from synthetic RGB data using the proposed method without any manual annotations, even in the source domain.

\begin{table}[t]
    \caption{Thermal segmentation performance before and after our adaptation using Intersection over Union (IoU).}
    \centering
    \resizebox{\columnwidth}{!}{
    \begin{tabular}{c|ccc}
    \hline
         & Non-water & Water & Average\\
    \hline
    Source only & 78.33 & 32.90 & 55.62 \\
    Our adapted model & 85.67 & 54.77 & 70.22 \\
    \hline
    \end{tabular}
    }
    \label{tab:water_seg}
\end{table}

\section{CONCLUSION}

This work presented an unsupervised RGB-to-thermal domain adaptation method using multi-domain attention network and adversarial attention learning, and demonstrated its effectiveness on both image classification and semantic segmentation tasks. Vision models adapted by our method achieved a large performance gain over source-only models, and performed on-par with supervised models trained on target. The proposed method can enable robots thermal vision ability without incurring the exorbitant costs of data labeling. In addition, our adaptation method is designed to keep the source performance, i.e. learn without forgetting, providing a unified vision model for both RGB and thermal images.

\bibliographystyle{bib/IEEEtran}
\bibliography{bib/strings-abrv, bib/IEEEabrv, bib/refs}

\end{document}